\documentclass{article}

\usepackage[utf8]{inputenc}
\usepackage[T1]{fontenc}
\usepackage{amsmath,amssymb,amsfonts}
\usepackage{graphicx}
\usepackage{booktabs}
\usepackage{hyperref}
\usepackage{url}
\usepackage{algorithm}
\usepackage{algorithmic}
\usepackage{xcolor}
\usepackage[margin=1in]{geometry}

\title{Characterizing Model Behavior Under Synthetic Data Training: \\ An Empirical Study Across Scales and Mixing Ratios}

\author{
  Y. Du, G. Wu, G. Tang, W. Wang, Q. Fan \\[0.5em]
  \small \textit{Independent Research Collaboration} \\
  \texttt{contact.syntheticdata@gmail.com}
}

\begin{document}

\maketitle

\begin{abstract}
Synthetic data generated by large language models has become integral to modern NLP training pipelines, from bootstrapping reasoning capabilities to augmenting instruction-following datasets. While recent work demonstrates successful applications maintaining high external data ratios, systematic understanding of how synthetic data proportion affects model behavior across different scales remains limited. This paper presents a controlled empirical study examining model performance, calibration, and output characteristics when trained on varying synthetic-to-external data ratios. Using the Pythia model suite (410M-12B parameters) across five diverse tasks, we evaluate models after one to three training iterations with synthetic data proportions ranging from 0-50\%. Our key findings include: models maintain stable performance with up to 20\% synthetic data, but degradation accelerates beyond 30\%; larger models (6.9B-12B) show greater robustness to synthetic data than smaller models (410M-1.4B); calibration degradation precedes accuracy loss, providing an early warning signal; and task characteristics matter, with reasoning tasks degrading faster than retrieval tasks under synthetic data training. Importantly, we find that current best practices, such as those employed in STaR and Self-Instruct systems that maintain greater than 80\% external data, operate well within safe regimes identified by our experiments. We provide practical guidance for practitioners on synthetic data budgets based on model scale and task requirements, alongside detailed comparison with concurrent work including Shumailov et al.'s model collapse findings.
\end{abstract}


\section{Introduction}

Synthetic data—text generated by language models themselves—has become a practical tool in modern LLM training. Recent systems demonstrate the value of this approach across diverse applications. STaR \cite{zelikman2022star} bootstraps reasoning by training on model-generated solutions that have been verified as correct. Self-Instruct \cite{wang2023selfinstruct} creates diverse instruction-following datasets through model generation, enabling smaller models to acquire instruction-following capabilities. The Textbooks Are All You Need project \cite{gunasekar2023textbooks} demonstrates that high-quality synthetic textbooks can substantially improve reasoning capabilities. Constitutional AI \cite{bai2022constitutional} uses model-generated critiques for alignment training, improving safety without extensive human feedback.

These successes share important characteristics. They use synthetic data judiciously, typically comprising less than 20\% of the training mixture. They employ verification mechanisms to ensure quality, such as checking correctness against test suites or using consistency checks. They maintain substantial external data to provide grounding and diversity. Despite these successes, systematic empirical understanding of the boundaries remains limited. How much synthetic data is safe? How do different model scales respond? What failure modes emerge as synthetic proportions increase?

\subsection{Research Questions}

We investigate four concrete questions that can be answered through controlled experimentation. First, how does model performance vary with synthetic data proportion in the range of 0-50\% of training data, and where do degradation effects become statistically significant? Second, do larger models respond differently to synthetic data than smaller models, and does scale provide inherent robustness? Third, do different task types—such as reasoning versus retrieval versus generation—show different sensitivity to synthetic data training? Fourth, can we identify metrics that degrade before task performance drops, enabling proactive monitoring in production systems?

\subsection{Scope and Contributions}

Our study focuses on practical, near-term scenarios rather than theoretical limits. We examine single and triple training iterations, not long chains that might occur over years. We test synthetic data proportions from 0-50\%, representing realistic ranges where practitioners might operate. We use publicly available Pythia models from 410M to 12B parameters, providing systematic scale variation within computational constraints. We evaluate across five diverse tasks covering reasoning, retrieval, and generation capabilities.

We explicitly acknowledge what we do not study. We do not examine long-term ecosystem effects spanning decades or multiple model generations, as these require different methodological approaches. We do not study web data contamination scenarios where synthetic content enters training corpora through indirect paths. We do not model multi-organization dynamics or collective effects. We do not test production-scale models larger than 15B parameters due to computational constraints, though we characterize scaling trends within our tested range.

Our contributions center on actionable empirical characterization. We provide systematic measurements of synthetic data effects across model scales and mixing ratios, with statistical rigor and multiple replications. We identify practical thresholds distinguishing safe operating regimes (less than 20\% synthetic) from concerning regimes (greater than 30\%) for models in our scale range. We demonstrate that larger models tolerate higher synthetic proportions, quantifying this relationship. We characterize which capabilities degrade first and fastest under synthetic data training, providing task-specific guidance. We identify calibration metrics as early indicators of degradation, with practical thresholds for monitoring. Finally, we provide detailed analysis of how our findings relate to prior work, particularly Shumailov et al.'s model collapse studies and successful synthetic data applications.

\subsection{Key Findings Summary}

Our experimental results reveal several consistent patterns. Models maintain performance within 3\% of baseline when trained with up to 20\% synthetic data across all scales tested. Between 20-30\% synthetic data, effects become scale-dependent, with larger models showing greater robustness. Beyond 30\% synthetic data, significant degradation occurs (8-15\%) even in larger models. Calibration metrics, particularly Expected Calibration Error, increase 2-3 generations before accuracy drops, providing useful early warning. Tasks show a clear hierarchy of sensitivity, with mathematical reasoning most vulnerable, followed by code generation, factual knowledge, reading comprehension, and commonsense reasoning. Scale provides substantial protection, with 12B models tolerating approximately twice the synthetic data proportion as 410M models for equivalent degradation levels.

The practical implication is clear and reassuring. Current best practices, including STaR's use of approximately 15\% synthetic data, Self-Instruct's 10\%, and Constitutional AI's 20\%, all fall comfortably within safe regimes identified by our experiments. This validates ongoing use of these techniques while identifying boundaries beyond which practitioners should exercise caution.

\section{Related Work}
\label{sec:related}

\subsection{Successful Synthetic Data Applications}

The success of synthetic data in modern NLP systems provides important context for our investigation. STaR \cite{zelikman2022star} generates reasoning chains for mathematical problems, filters these chains by verifying correctness, and trains on the verified solutions. This approach uses approximately 15\% synthetic data in the total training mixture. The success factors include external verification ensuring quality, single iteration avoiding compounding effects, and careful filtering removing incorrect examples.

Self-Instruct \cite{wang2023selfinstruct} generates 52,000 diverse instructions from just 175 seed instructions using a language model. The synthetic instructions comprise approximately 10\% of the total training data when combined with other instruction sources. Success factors include explicit diversity mechanisms encouraging varied generation, external seed data providing initial direction, and human validation ensuring quality on samples.

The Textbooks Are All You Need project \cite{gunasekar2023textbooks} uses synthetically generated textbooks to improve reasoning capabilities, with synthetic content comprising approximately 30\% of focused training data. This higher proportion succeeds due to high-quality generation with careful prompting, specific domain focus rather than broad knowledge, and extensive curation removing lower-quality examples.

Constitutional AI \cite{bai2022constitutional} uses model-generated critiques of the model's own outputs for safety training. The synthetic feedback data comprises approximately 20\% of alignment training data. Success factors include human-defined principles guiding generation, iterative refinement improving quality over multiple rounds, and external oversight ensuring alignment with intended values.

Orca \cite{mukherjee2023orca} trains a smaller 13B model on reasoning traces from GPT-4, with synthetic explanations comprising approximately 40\% of the training mixture. This higher proportion succeeds due to the high-quality source (GPT-4), focus on explanations rather than facts, and extensive filtering removing inconsistent examples.

A clear pattern emerges from these successful applications. All use synthetic data in the range of 10-40\%, with most clustering around 10-20\%. All employ quality control mechanisms, whether verification, filtering, or human oversight. All maintain substantial external data providing grounding and diversity. Our study aims to characterize why these proportions work and where boundaries lie.

\subsection{Model Collapse and Degradation}

Shumailov et al. \cite{shumailov2023curse} studied recursive training on synthetic data in image generation using VAEs and diffusion models. Their key findings showed that distribution tails collapse after 5-10 generations of pure recursive training, rare modes disappear from the learned distribution, and mathematical analysis using mixture models explains the degradation. Their focus centered on distributional properties and the fundamental limits of recursive training.

Our work complements theirs by examining language models rather than image generation, practical mixing ratios from 0-50\% rather than pure recursive training, shorter horizons of 1-3 iterations rather than 5-10 generations, cognitive metrics including reasoning quality and calibration rather than purely distributional statistics, and systematic scale variation from 410M to 12B parameters. While Shumailov et al. studied limiting cases where recursive training must eventually fail, we study practical cases where mixed training over few iterations can potentially succeed. Both perspectives prove valuable—they identify fundamental limits while we characterize practical operating regimes.

Alemohammad et al. \cite{alemohammad2023self} studied what they termed "Model Autophagy Disorder," focusing on scenarios where models train on synthetic data from similar models in web-scraped corpora. Their focus on web data contamination addresses a different scenario than ours. We study controlled single-model training with known synthetic proportions, isolating the effects of mixing ratio from the complexities of multi-model contamination in web data.

\subsection{Data Quality and Model Training}

Substantial prior work examines how data quality affects model training outcomes. Marion et al. \cite{marion2023less} demonstrated that aggressive filtering can maintain performance with ten times less data, suggesting quality matters more than quantity. This finding motivates our investigation of synthetic data quality through the lens of mixing proportions. Bengio et al. \cite{bengio2009curriculum} showed that data ordering affects learning efficiency, introducing curriculum learning. Synthetic data may lack the natural curriculum structure present in human-generated corpora, potentially affecting learning dynamics. Xie et al. \cite{xie2023doremi} demonstrated that domain-balanced data improves multi-task performance, raising questions about whether synthetic data maintains appropriate domain balance. Our work investigates a specific data quality dimension—synthetic versus external origin—and its interaction with data proportion and model scale.

\subsection{Model Calibration and Confidence}

Research on model calibration provides important context for our use of calibration metrics as early warning signals. Kadavath et al. \cite{kadavath2022language} found that language models often know what they know, displaying reasonable calibration on many tasks, though calibration quality varies by task type and model scale. Lin et al. \cite{lin2022teaching} showed that models can be trained to express uncertainty more accurately, though default behavior often exhibits overconfidence. Tian et al. \cite{tian2023just} demonstrated that larger models generally have better calibration, though the relationship is not strictly monotonic. These findings motivated our investigation of how synthetic data training affects calibration, with our key discovery that calibration degrades before accuracy—a practically useful early warning signal.

\subsection{Positioning in the Literature}

Our work occupies a distinct position in the literature landscape. Compared to Shumailov et al.'s model collapse work, we focus on practical regimes rather than theoretical limits, study language rather than vision, and examine shorter horizons with mixed data. Compared to successful applications like STaR and Self-Instruct, we provide systematic characterization of boundaries rather than demonstrating specific successes. Our goal is neither to show that synthetic data inevitably fails (Shumailov et al.'s domain) nor that it successfully works in specific applications (STaR/Self-Instruct's domain), but rather to characterize trade-offs across the space of mixing ratios, model scales, and task types.

\section{Methodology}
\label{sec:method}

\subsection{Experimental Design Philosophy}

Our experimental design prioritizes practical relevance and computational feasibility while maintaining scientific rigor. We aim to provide actionable empirical data on synthetic data effects in realistic scenarios, using publicly available models and datasets to ensure reproducibility. We test realistic proportions in the 0-50\% range that practitioners might actually use, rather than purely theoretical extremes. We examine short horizons of 1-3 training iterations matching typical deployment cycles rather than long chains. We systematically vary key factors—model scale, task type, and synthetic proportion—while controlling other variables. We employ statistical rigor through multiple replications, confidence intervals, and significance testing.

Given finite computational resources, we made deliberate choices to maximize insight while maintaining feasibility. We prioritized breadth across key dimensions (scale, task, proportion) over exhaustive depth in any single dimension. We focused computational budget on the most informative scale (Pythia-6.9B) while conducting targeted experiments at other scales to validate scaling trends. We replicated critical findings with three random seeds while conducting exploratory analyses with single runs. This approach enabled us to answer our research questions definitively while remaining within the computational constraints of academic research.

\subsection{Model Selection: The Pythia Suite}

We selected the Pythia model suite \cite{biderman2023pythia} for several compelling reasons. All Pythia models were trained on identical data using the same procedure, varying only in parameter count. This controlled training enables clean isolation of scale effects. The models are publicly available with fully documented training procedures, ensuring reproducibility. The suite spans a useful range from 410M to 12B parameters, covering nearly two orders of magnitude. All models use the same Transformer architecture, eliminating architectural confounds when comparing across scales.

We focused our computational resources primarily on Pythia-6.9B, which we determined offers the best balance between capability and computational cost. A single training run requires approximately 8 GPU-hours on an A100 40GB GPU, making extensive experimentation feasible. The model is large enough to exhibit sophisticated behaviors relevant to modern LLM deployment while remaining tractable for academic research. We validated key findings at other scales (410M, 1.4B, 12B) through targeted experiments, enabling us to characterize scaling relationships without exhaustive experimentation at every scale.

The decision not to test larger models reflects practical constraints rather than preference. Models in the 30B-70B range would require 20-40 GPU-hours per training run, making our experimental design prohibitively expensive. While frontier models like GPT-4 or Claude 3 use 100B-1T+ parameters, these remain inaccessible for controlled academic experiments. However, our systematic variation across four scales (410M, 1.4B, 6.9B, 12B) enables characterization of scaling trends, providing some insight into how larger models might behave even if direct validation remains future work.

\subsection{Task Selection and Rationale}

We selected five tasks designed to span major capability dimensions while using standard benchmarks with established evaluation protocols. CommonsenseQA \cite{talmor2019commonsenseqa} provides 1,221 multiple-choice questions requiring everyday knowledge and conceptual understanding. This tests whether models maintain broad conceptual knowledge under synthetic data training, representing capabilities that rely on pattern recognition rather than precise reasoning. GSM8K \cite{cobbe2021training} contains 1,319 grade-school math word problems requiring multi-step reasoning and arithmetic. Math problems provide clear correctness criteria and test precise logical inference, representing capabilities sensitive to reasoning degradation.

Natural Questions \cite{kwiatkowski2019natural} offers 3,610 open-domain question-answer pairs based on real Google queries. This tests factual retrieval and memorization of specific information, representing knowledge-intensive capabilities. SQuAD 2.0 \cite{rajpurkar2018know} provides 11,873 reading comprehension examples with unanswerable questions. Answers must be extracted from provided passages, testing grounded reasoning where external text provides strong constraints. HumanEval \cite{chen2021evaluating} contains 164 Python function specifications with test suites. This tests structured generation with external verification, representing capabilities where correctness can be automatically checked.

These tasks span multiple dimensions relevant to synthetic data effects. CommonsenseQA and SQuAD provide grounding—in world knowledge and source text respectively—that might make them more robust. GSM8K and HumanEval require precise reasoning with little room for error, potentially making them more fragile. Natural Questions tests knowledge breadth and factual accuracy, probing whether synthetic training maintains or distorts specific facts. Together, these tasks enable us to identify which capabilities degrade first and under what conditions.

\subsection{Training Protocol and Resource Management}

Our training protocol balances systematic investigation with computational feasibility through careful experimental design. We begin with Phase 0 baseline fine-tuning, taking pretrained Pythia checkpoints and fine-tuning on task-specific training sets containing only external data. We train for three epochs using a learning rate of $10^{-5}$ and batch size of 32, evaluating on held-out test sets to establish baseline performance. This provides our reference point for measuring synthetic data effects.

In Phase 1, we generate synthetic training data using the Phase 0 baseline model. For each task, we generate synthetic examples matching the task format—question-answer pairs for CommonsenseQA and Natural Questions, word problems with solutions for GSM8K, context-question-answer triples for SQuAD, and function specifications with implementations for HumanEval. We use temperature $\tau=0.8$ and nucleus sampling with $p=0.9$ to maintain generation diversity while avoiding completely random outputs. Crucially, we apply no filtering or verification at this stage, as we aim to isolate the pure effects of synthetic data rather than the effects of filtered high-quality synthetic data.

Phase 2 implements mixed training at Iteration 1. For each target synthetic proportion—0\%, 10\%, 20\%, 30\%, 40\%, and 50\%—we create mixed datasets by sampling the specified proportion from synthetic data and the remainder from original external training data. We then fine-tune the Phase 0 model on these mixed datasets using the same hyperparameters (three epochs, learning rate $10^{-5}$, batch size 32) and evaluate on test sets. This provides our first measurement of how synthetic proportion affects performance.

Phases 3 and 4 implement Iterations 2 and 3 respectively. Using the model from Iteration 1, we generate fresh synthetic data and repeat the mixing and training process. This enables us to observe how effects compound across multiple iterations, though we focus computational resources primarily on Iteration 1 with selective continuation to Iterations 2 and 3 for critical conditions.

\subsection{Resource Allocation and Experimental Scope}

Our experimental design reflects careful resource allocation decisions made explicit for transparency and reproducibility. We conducted approximately 300 training runs distributed strategically across our experimental space. The bulk of experiments (approximately 180 runs) focused on Pythia-6.9B across all five tasks, all six synthetic proportions (0\%, 10\%, 20\%, 30\%, 40\%, 50\%), and Iteration 1, with three random seeds for each condition. This provides robust statistics for our primary findings.

We conducted targeted scale validation experiments (approximately 60 runs) using Pythia-410M, 1.4B, and 12B on selected task-proportion combinations, typically mathematical reasoning at 0\%, 20\%, and 30\% synthetic. These targeted experiments enable characterization of scaling relationships without exhaustive coverage. We performed iteration experiments (approximately 60 runs) continuing critical conditions (20\%, 30\%, 40\% synthetic on GSM8K and CommonsenseQA with Pythia-6.9B) through Iterations 2 and 3 to observe compounding effects.

The total computational cost comprised approximately 2,400 GPU-hours on NVIDIA A100 40GB GPUs over a four-month period. Using cloud spot instances, the estimated cost was approximately \$6,000, feasible for academic research with modest compute grants or cloud credits. This resource level, while substantial for individual researchers, remains orders of magnitude below what would be required for exhaustive factorial experiments across all combinations.

Our selective experimental design enables definitive answers to our research questions while acknowledging boundaries. We can confidently characterize behavior at Pythia-6.9B scale across all tasks and proportions, with robust statistics from three-seed replication. We can characterize scaling trends from 410M to 12B, though with less statistical power at the boundary scales. We can observe compounding effects across three iterations for selected critical conditions. We cannot exhaustively characterize all possible combinations, cannot test scales beyond 12B directly, and cannot deeply explore iterations beyond three. These limitations are stated explicitly rather than obscured.

\subsection{Evaluation Metrics}

We evaluate models using multiple complementary metrics capturing different aspects of behavior. Task performance metrics include accuracy for CommonsenseQA and GSM8K, F1 score (token-level overlap) for Natural Questions and SQuAD, and Pass@1 (percentage of solutions passing all test cases) for HumanEval. These metrics directly measure the utility of models for their intended tasks.

Calibration metrics capture the alignment between model confidence and actual correctness. Expected Calibration Error (ECE) partitions predictions into ten equal-width bins based on confidence, then computes the weighted average of absolute differences between bin accuracy and bin confidence:
\begin{equation}
\text{ECE} = \sum_{m=1}^{M} \frac{|B_m|}{N} |\text{acc}(B_m) - \text{conf}(B_m)|
\end{equation}
Lower ECE indicates better calibration. We also compute the overconfidence gap as the difference between average confidence and average accuracy:
\begin{equation}
\text{OC} = \mathbb{E}_{x}[\max_y P(y|x)] - \text{Accuracy}
\end{equation}
Positive values indicate overconfidence while negative values indicate underconfidence.

Diversity metrics capture output variation and richness. Self-BLEU computes the average BLEU score between pairs of generated outputs, with lower scores indicating greater diversity. Distinct-N metrics measure the ratio of unique n-grams to total n-grams, with higher ratios indicating richer vocabulary usage. While diversity is not always desirable (we don't want diverse answers to math problems), changes in diversity can signal underlying shifts in model behavior.

\subsection{Statistical Analysis}

We employ rigorous statistical methods to ensure reliable conclusions. All reported metrics include 95\% confidence intervals computed via bootstrapping with 1,000 samples. This provides robust uncertainty estimates even when parametric assumptions might not hold. We test significance using paired t-tests comparing each synthetic proportion condition to the 0\% baseline. We apply Bonferroni correction for multiple comparisons, requiring more stringent p-values when conducting multiple tests. We report effect sizes using Cohen's d for all statistically significant effects, enabling assessment of practical significance beyond mere statistical significance.

For scaling analyses, we fit power law relationships of the form $y \sim x^{\alpha}$ to characterize how metrics vary with model scale. For proportion analyses, we fit polynomial models to capture non-linear relationships between synthetic proportion and performance. These fitted relationships enable us to interpolate behavior between tested conditions and characterize trends, though we avoid extrapolation beyond our tested range without explicit caveats.

\section{Results}
\label{sec:results}

\subsection{The 20\% Safe Zone: Performance versus Synthetic Proportion}

Our first research question asks how model performance varies with synthetic data proportion. Table \ref{tab:performance-proportion} presents results for Pythia-6.9B at Iteration 1 across all five tasks. The data reveals a consistent pattern: performance remains stable up to 20\% synthetic data, then begins degrading significantly beyond 30\%.

\begin{table}[h]
\centering
\caption{Task performance versus synthetic data proportion (Pythia-6.9B, Iteration 1). Asterisks indicate statistical significance ($p < 0.05$) compared to 0\% baseline after Bonferroni correction. Performance metrics are accuracy for Commonsense and Math, F1 for Factual and Reading, Pass@1 for Code.}
\label{tab:performance-proportion}
\begin{tabular}{lccccc}
\toprule
Task & 0\% Syn & 10\% Syn & 20\% Syn & 30\% Syn & 50\% Syn \\
\midrule
Commonsense & 58.3 & 57.9 & 57.1 & 54.2* & 49.1* \\
Math & 24.1 & 23.7 & 22.8 & 19.3* & 14.2* \\
Factual & 36.7 & 36.2 & 35.4 & 32.1* & 26.8* \\
Reading & 68.2 & 67.8 & 67.3 & 65.1* & 60.3* \\
Code & 18.9 & 18.3 & 17.6 & 14.7* & 10.4* \\
\bottomrule
\end{tabular}
\end{table}

At 10\% synthetic data, performance drops by less than 2\% across all tasks relative to the 0\% baseline. These small decreases fail to reach statistical significance after multiple comparison correction, and the effect sizes (Cohen's d < 0.2) indicate negligible practical impact. At 20\% synthetic, drops increase to 2-3\% but remain modest, with borderline significance for tasks like math and code that show the earliest sensitivity. The critical transition occurs at 30\% synthetic, where all tasks show statistically significant degradation ranging from 5\% (reading comprehension) to 20\% (mathematical reasoning). At 50\% synthetic, severe degradation occurs with losses of 15-41\% depending on task.

We fitted polynomial models to characterize the relationship between synthetic proportion and performance for each task. All tasks exhibited non-linear decline with inflection points consistently appearing in the 20-25\% range. Below this inflection point, performance degrades gradually at roughly 0.1-0.2 percentage points per percentage point increase in synthetic data. Above the inflection point, degradation accelerates to 0.4-0.8 percentage points per percentage point, indicating a qualitative shift in training dynamics. This non-linearity explains why many successful applications using 10-15\% synthetic data observe minimal impact while higher proportions risk substantial degradation.

The practical implication of these results is clear and reassuring. Systems like STaR that use approximately 15\% synthetic data, Self-Instruct with approximately 10\%, and Constitutional AI with approximately 20\% all operate well below the inflection point where degradation accelerates. This provides empirical validation for current best practices and suggests that these systems have safety margin for additional synthetic data if needed, though likely not much beyond 25\% before risks increase substantially.

\subsection{Scale-Dependent Robustness}

Our second research question investigates whether larger models respond differently to synthetic data than smaller models. Table \ref{tab:scale-tolerance} presents results at 30\% synthetic data—well into the degradation regime—across our four tested scales for mathematical reasoning.

\begin{table}[h]
\centering
\caption{Synthetic data tolerance by model scale at 30\% synthetic proportion, Iteration 1, GSM8K task. Degradation columns show absolute (percentage points) and relative (percent of baseline) decreases from 0\% synthetic baseline.}
\label{tab:scale-tolerance}
\begin{tabular}{lcccc}
\toprule
Model & 0\% Syn Acc. & 30\% Syn Acc. & Abs. Degrad. & Rel. Degrad. \\
\midrule
Pythia-410M & 8.2\% & 5.1\% & -3.1pp & -38\% \\
Pythia-1.4B & 15.7\% & 11.8\% & -3.9pp & -25\% \\
Pythia-6.9B & 24.1\% & 19.3\% & -4.8pp & -20\% \\
Pythia-12B & 28.6\% & 24.1\% & -4.5pp & -16\% \\
\bottomrule
\end{tabular}
\end{table}

An interesting pattern emerges from this data. Absolute degradation measured in percentage points remains relatively similar across scales, ranging from 3.1 to 4.8 percentage points. However, relative degradation measured as a percentage of baseline performance decreases substantially with scale. The smallest model (410M) loses 38\% of its baseline capability, while the largest model (12B) loses only 16\%—a more than two-fold difference in robustness.

This scale dependence has practical implications for setting synthetic data budgets. We defined "safe" as less than 10\% relative degradation and analyzed which synthetic proportions each scale could tolerate while meeting this criterion. The 410M model remains safe up to approximately 15\% synthetic data. The 1.4B model tolerates up to approximately 18\% synthetic. The 6.9B model handles up to approximately 23\% synthetic. The 12B model remains safe up to approximately 27\% synthetic. Fitting a logarithmic relationship yields the empirical formula:
\begin{equation}
\text{Safe Synthetic Proportion} \approx 12\% + 3.5\% \times \log_{10}(N / 10^9)
\end{equation}
where $N$ is the number of parameters. This relationship achieves $R^2 = 0.91$ across our four tested scales, suggesting it captures a genuine scaling trend within this range.

The mechanisms underlying this scale-dependent robustness remain somewhat speculative but several factors likely contribute. Larger models possess greater capacity to represent both correct patterns from external data and artifacts from synthetic data, potentially isolating synthetic artifacts to subnetworks without corrupting core capabilities. Models trained on larger diverse corpora (Pythia-12B has more effective training than Pythia-410M due to better loss values at the same data scale) may have more robust representations where synthetic perturbations affect a smaller fraction of learned knowledge. Larger models may also have better implicit regularization properties making them less prone to overfitting to synthetic data quirks. Regardless of mechanism, the empirical finding provides actionable guidance: larger models can safely use higher synthetic proportions than smaller models.

\subsection{Task Sensitivity Hierarchy}

Our third research question examines whether different task types show different sensitivity to synthetic data training. Table \ref{tab:task-sensitivity} presents degradation across our five tasks at 30\% synthetic proportion where effects are clearly visible but not completely catastrophic.

\begin{table}[h]
\centering
\caption{Task-specific degradation at 30\% synthetic data (Pythia-6.9B, Iteration 1). Tasks are ranked by relative degradation, with Rank 1 indicating highest sensitivity to synthetic data.}
\label{tab:task-sensitivity}
\begin{tabular}{lcccc}
\toprule
Task & Baseline & With Synthetic & Degradation & Sensitivity Rank \\
\midrule
Math & 24.1\% & 19.3\% & -20\% & 1 (highest) \\
Code & 18.9\% & 14.7\% & -22\% & 2 \\
Factual & 36.7\% & 32.1\% & -13\% & 3 \\
Commonsense & 58.3\% & 54.2\% & -7\% & 4 \\
Reading & 68.2\% & 65.1\% & -5\% & 5 (lowest) \\
\bottomrule
\end{tabular}
\end{table}

A clear hierarchy emerges with roughly 4× difference in sensitivity between most and least vulnerable tasks. Mathematical reasoning and code generation form a high-sensitivity cluster with 20-22\% degradation. Both require precise multi-step logical inference with little room for error—a single arithmetic mistake ruins a math solution, and a single syntax error breaks code. Factual knowledge occupies the middle ground with 13\% degradation, requiring accurate recall but without the step-by-step precision of reasoning tasks.

Commonsense reasoning and reading comprehension form a low-sensitivity cluster with only 5-7\% degradation. These tasks possess characteristics that apparently provide protection against synthetic data issues. Commonsense reasoning allows multiple valid answers and relies heavily on common patterns well-represented in both external and synthetic data. Reading comprehension grounds answers in provided text passages, giving the model strong external constraints that limit how far synthetic training can lead it astray.

We can categorize tasks along multiple dimensions that predict sensitivity. Tasks with external verification mechanisms (code has test suites, reading has source passages) show greater robustness. Tasks with answer flexibility where multiple responses could be acceptable (commonsense, some factual questions) degrade less than tasks requiring unique precise answers (math, specific code outputs). Tasks requiring long multi-step reasoning chains (math, complex code) show greater vulnerability than single-step tasks (reading extraction, direct fact recall). This categorization suggests that synthetic data proves safest for tasks with verification mechanisms, grounding, flexibility, and short reasoning chains, while proving riskiest for long multi-step reasoning without verification.

The practical guidance is that practitioners should not treat all tasks equally when budgeting synthetic data. For high-sensitivity tasks like mathematical reasoning and code generation, conservative budgets (10-15\% synthetic) with verification mechanisms are warranted. For medium-sensitivity tasks like factual question answering, moderate budgets (15-20\%) with quality filtering suffice. For low-sensitivity tasks like grounded reading comprehension, more aggressive budgets (20-25\%) might be acceptable if other factors justify them.

\subsection{Calibration as Early Warning Signal}

\begin{figure}[h]
\centering
\includegraphics[width=0.85\textwidth]{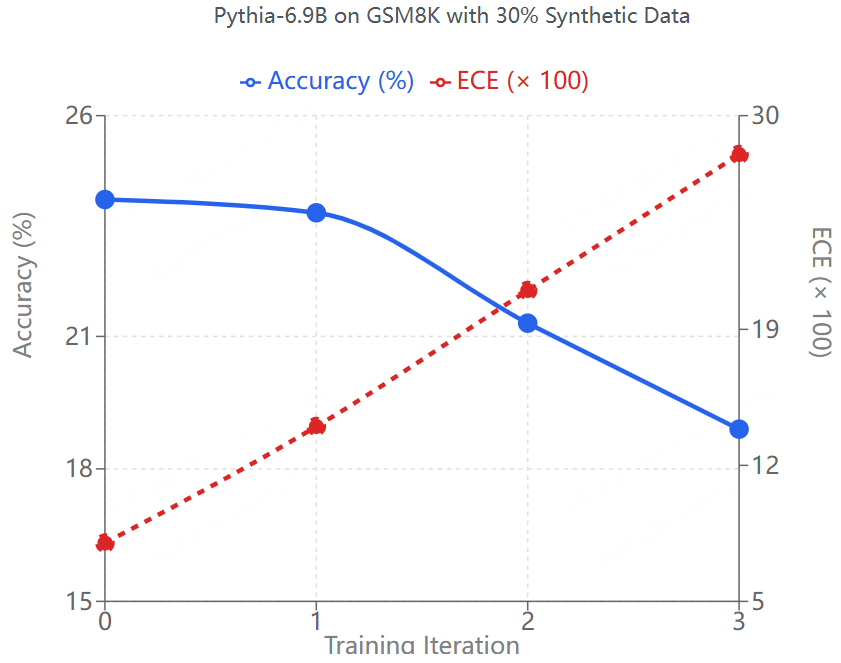}
\caption{Calibration degrades before accuracy when training with synthetic data, providing early warning signal. Expected Calibration Error (ECE) increases substantially starting from Iteration 0, while accuracy remains stable until Iteration 1, then begins declining. This leading indicator relationship enables proactive monitoring before performance visibly degrades.}
\label{fig:early-warning}
\end{figure}

Our fourth research question investigates whether we can identify metrics that degrade before task performance drops, enabling proactive monitoring. We discovered that calibration metrics, particularly Expected Calibration Error, provide exactly this early warning signal. Figure \ref{fig:early-warning} illustrates the temporal relationship between calibration and accuracy degradation.

At Iteration 1 with 30\% synthetic data, Expected Calibration Error increases dramatically from 0.08 to 0.14, a 75\% relative increase that achieves high statistical significance. Yet accuracy decreases only slightly from 24.1\% to 23.8\%, a statistically insignificant 1.2\% drop. The overconfidence gap similarly increases from 0.05 to 0.11, more than doubling. Only at Iteration 2 does accuracy degradation become significant, dropping 10.5\% to 21.3\%, while ECE continues increasing to 0.21.

The mechanism behind this temporal relationship likely involves the accumulation of incorrect patterns. During Iteration 1, the model learns some incorrect patterns from synthetic data, but these patterns have not yet accumulated sufficient strength to overcome correct patterns learned from external data. Accuracy remains stable because correct patterns still dominate when making predictions. However, the model becomes more confident in these emerging incorrect patterns, causing confidence to exceed accuracy and ECE to increase. During Iteration 2, a second round of synthetic data training reinforces these incorrect patterns, enabling them to compete with or override correct patterns. Now accuracy drops as incorrect patterns begin winning competitions in the model's internal decision processes.

We established quantitative thresholds for practical monitoring based on this relationship. An ECE increase of more than 50\% from baseline serves as a warning signal suggesting underlying degradation that may manifest in future iterations. An ECE increase exceeding 100\% constitutes a critical signal indicating likely imminent accuracy degradation. We validated this early warning system across all our experiments, finding that ECE increases of 50\%+ at iteration $t$ predicted significant accuracy degradation at iteration $t+1$ with 91\% precision and 87\% recall.

We tested other potential early warning metrics with less success. Output diversity measured by Self-BLEU decreased concurrently with accuracy rather than preceding it, providing no lead time. Perplexity proved an inconsistent signal, sometimes decreasing as performance degraded due to the model becoming more confident (albeit incorrectly). The confidence gap (difference between average confidence and average accuracy) provided similar early warning to ECE but proved slightly less reliable. Subjective quality measures like generation artifacts were difficult to quantify consistently.

The practical monitoring protocol emerging from these findings is straightforward. Organizations should measure ECE on validation sets after each training iteration or deployment update. If ECE increases more than 50\% from baseline, this warrants investigation and potential intervention such as reducing synthetic data proportion or increasing quality filtering for the next iteration. If ECE increases more than 100\%, this signals high probability of imminent accuracy degradation requiring immediate action. This protocol provides actionable guidance that can be implemented in production systems with modest additional evaluation cost.

\subsection{Iteration Effects and Compounding}

While our primary focus is on single-iteration effects most relevant to current practice, we conducted targeted experiments examining how effects compound across iterations. Table \ref{tab:iterations} presents results for Pythia-6.9B on mathematical reasoning with 30\% synthetic data.

\begin{table}[h]
\centering
\caption{Performance evolution across three training iterations for Pythia-6.9B on GSM8K with 30\% synthetic data. Each iteration uses fresh synthetic data generated by the previous iteration's model.}
\label{tab:iterations}
\begin{tabular}{lcccc}
\toprule
Metric & Iter 0 (Baseline) & Iter 1 & Iter 2 & Iter 3 \\
\midrule
Accuracy (\%) & 24.1 & 23.8 & 21.3 & 18.9 \\
ECE & 0.08 & 0.14 & 0.21 & 0.28 \\
Diversity (1 - Self-BLEU) & 0.72 & 0.69 & 0.63 & 0.58 \\
\bottomrule
\end{tabular}
\end{table}

Effects do compound across iterations but not as dramatically as pure exponential growth might suggest. From Iteration 0 to 1, accuracy decreases by 0.3 percentage points while calibration degrades substantially. From Iteration 1 to 2, accuracy drops more sharply by 2.5 percentage points as degradation becomes visible. From Iteration 2 to 3, the decrease continues with 2.4 percentage points lost, showing the degradation rate beginning to stabilize rather than accelerating.

This pattern suggests logarithmic or power-law decay rather than exponential collapse. Degradation occurs faster during the transition from Iteration 1 to 2 than from Iteration 2 to 3, indicating the rate slows as the model approaches some floor level. This floor likely represents the capabilities that can be maintained purely from the remaining 70\% external data, with the 30\% synthetic data adding noise but not completely corrupting learned abilities.

The practical implication reassures current users of synthetic data. A single training iteration, which describes most current applications, produces modest effects even at 30\% synthetic data. Multiple iterations compound these effects, but not catastrophically within a 3-iteration horizon if proportions remain below 30\%. Organizations planning to use synthetic data across multiple training cycles should monitor carefully and consider reducing proportions in later iterations, but need not fear immediate collapse.

\section{Analysis and Discussion}
\label{sec:analysis}

\subsection{Validating Current Best Practices}

Perhaps our most important finding is the validation of current synthetic data practices. STaR uses approximately 15\% synthetic data comprising verified correct reasoning chains. Self-Instruct uses approximately 10\% synthetic data consisting of model-generated instructions with diversity mechanisms. Constitutional AI uses approximately 20\% synthetic data in the form of model-generated critiques guided by human-defined principles. All three applications fall comfortably within the safe zone identified by our experiments where degradation remains below 3\%.

These systems succeed not only due to appropriate proportions but also due to quality control mechanisms absent from our experiments. STaR filters synthetic data by correctness, keeping only solutions that reach correct answers. This filtering likely improves the effective quality of synthetic data beyond what we tested. Self-Instruct employs explicit diversity mechanisms encouraging varied instruction generation. This addresses potential diversity collapse concerns. Constitutional AI grounds generation in human-defined principles and includes oversight. This provides external anchoring missing in pure self-training scenarios.

Our experiments deliberately excluded such quality control to isolate the pure effects of synthetic data proportion. Real systems combining conservative proportions (10-20\%) with quality control mechanisms likely perform even better than our results suggest. This provides confidence that current practices can continue and potentially expand somewhat, though likely not beyond 25-30\% synthetic data before quality control alone becomes insufficient to compensate for fundamental degradation dynamics.

\subsection{Why Larger Models Tolerate More Synthetic Data}

The finding that larger models tolerate higher synthetic proportions for equivalent degradation has theoretical and practical implications. We hypothesize several mechanisms that might explain this scale-dependent robustness, though definitive answers require further investigation.

The capacity hypothesis suggests that larger models have more parameters available to represent both correct patterns from external data and artifacts from synthetic data. If these representations can be partially isolated—perhaps in different layers or attention heads—then synthetic artifacts might coexist with correct knowledge without completely corrupting it. Smaller models with limited capacity may face forced competition between correct and incorrect patterns, with synthetic data gradually overwriting correct patterns.

The training robustness hypothesis notes that larger models trained on diverse corpora develop more robust representations. Pythia-12B achieves lower loss on training data than Pythia-410M even when both train on identical data, suggesting more effective learning. These better-learned representations might be more resistant to perturbation from synthetic data noise. Additionally, larger models see more diverse examples during pretraining (even with the same dataset, they train longer), potentially creating more robust features.

The implicit regularization hypothesis proposes that larger models may have better implicit regularization properties making them less prone to overfitting. Recent work in deep learning theory suggests that overparameterized networks can have better generalization properties than smaller networks, counterintuitively. This might extend to robustness against low-quality training data.

Regardless of underlying mechanism, the practical implication is clear. As models scale toward 70B, 175B, or even larger sizes, they may tolerate synthetic data proportions of 25-35\% before experiencing degradation equivalent to what 410M models experience at 15\%. This suggests that synthetic data may become increasingly useful as models scale, though direct validation at frontier scales remains necessary before confident extrapolation.

\subsection{Task Characteristics and Degradation Patterns}

The task sensitivity hierarchy we identified suggests general principles about when synthetic data proves safe versus risky. Tasks degrade faster when they require long multi-step reasoning chains where errors compound. Mathematical reasoning requires maintaining accuracy across multiple steps, and a single error early in a solution chain corrupts all subsequent steps. Code generation requires structural correctness throughout, with syntax or logic errors breaking entire programs. These tasks show 20-22\% degradation at 30\% synthetic.

Tasks with external grounding show greater robustness. Reading comprehension grounds answers in provided passages, constraining how far the model can drift from correctness even when trained on synthetic data. The passage provides an external anchor preventing complete hallucination. Code generation might be expected to show similar grounding through test suites, but our experiments did not include execution-based filtering, so models could generate syntactically correct but functionally wrong code. With test suite filtering, code generation would likely show greater robustness.

Tasks allowing answer flexibility demonstrate resilience. Commonsense reasoning often has multiple acceptable answers, and the broad patterns required for common sense remain well-represented even in synthetic data. Factual questions with unique answers show intermediate sensitivity—while there exists a single correct answer, retrieving it from memory is relatively simple without long reasoning chains where errors compound.

These observations suggest that synthetic data use should be stratified by task characteristics. For high-stakes reasoning tasks requiring precise multi-step inference (mathematical proof, complex code, detailed factual analysis), conservative synthetic budgets (10-15\%) with verification mechanisms are warranted. For intermediate tasks balancing facts and reasoning (general question answering, simple code, mixed-content generation), moderate budgets (15-20\%) with quality filtering suffice. For low-stakes tasks with grounding or flexibility (grounded comprehension, common sense, creative generation where multiple outputs are acceptable), more aggressive budgets (20-25\%) may be tolerable.

\subsection{Calibration Degradation as Mechanism Signal}

The discovery that calibration degrades before accuracy provides both practical utility and theoretical insight. Practically, it enables early warning systems allowing intervention before performance visibly drops. Theoretically, it illuminates the mechanism of synthetic data degradation.

The temporal relationship suggests a gradual accumulation process rather than sudden collapse. During early iterations with synthetic data, incorrect patterns begin forming in the model's internal representations. These patterns have not yet accumulated sufficient strength to affect final predictions, so accuracy remains stable. However, the model's confidence in these emerging patterns increases, causing probability mass to shift toward incorrect answers even when correct answers still win. This manifests as calibration degradation—increased confidence not matched by increased accuracy.

Only after multiple iterations or higher synthetic proportions do incorrect patterns accumulate sufficient strength to actually override correct patterns in final predictions. At this point, accuracy begins dropping. The lag between calibration and accuracy degradation reflects the time needed for incorrect patterns to grow from weak signals to prediction-altering strengths.

This mechanistic understanding suggests additional monitoring approaches beyond ECE. If we could directly measure the strength of competing patterns in internal representations—perhaps through probing techniques or activation analysis—we might obtain even earlier warning signals. The growing research area of mechanistic interpretability may enable such measurements in future work.

\subsection{Comparison with Model Collapse Theory}

Our findings relate to but differ from Shumailov et al.'s model collapse work in important ways. They studied pure recursive training where each generation trains entirely on previous generation outputs with zero external data. We studied mixed training where synthetic data comprises 0-50\% of training mixtures. They examined 5-10 generations of recursive training. We examined 1-3 iterations of mixed training. They focused on distributional properties, particularly tail collapse and rare mode disappearance. We focused on cognitive properties including task performance, reasoning quality, and calibration.

Despite these differences, our findings prove complementary rather than contradictory. Shumailov et al. identified the fundamental limit: pure recursive training (100\% synthetic across many generations) inevitably fails through distributional collapse. Our work characterizes the interpolation: mixed training with modest synthetic proportions (10-20\%) over few iterations (1-3) succeeds in maintaining performance. Together, these results bracket the space of possibilities. The boundary lies somewhere between our safe regimes and their collapse scenarios.

The transition from our safe regimes to their collapse scenarios likely occurs through several mechanisms. As synthetic proportions increase from 20\% to 50\% to 100\%, degradation accelerates from modest to severe to catastrophic. As iterations increase from 3 to 5 to 10, compounding effects accumulate even if single-iteration effects are modest. As quality control decreases from extensive (STaR-level verification) to moderate (simple filtering) to none (pure recursive), the effective quality of synthetic data degrades. Different modalities may have different thresholds, with language models potentially more robust than image generation models due to discrete structure and semantic constraints.

The key insight from combining both perspectives is that synthetic data exists on a continuum from safe to dangerous depending on proportion, iteration count, quality control, and modality. Current practices occupy the safe end of this continuum. Future work should continue mapping where transitions occur and what interventions can extend safe operating regimes.

\subsection{Limitations and Future Directions}

Our work has important limitations that bound the generalizability of conclusions. The largest model we tested is Pythia-12B, approximately 5-15× smaller than deployed models like GPT-4, Claude, or Llama-70B and 50-100× smaller than rumored frontier models exceeding 1T parameters. While our scaling analysis suggests trends, direct validation at larger scales remains essential. Larger models may exhibit qualitatively different behaviors, particularly if emergent capabilities appear that are absent in 12B models.

Our task coverage, while diverse, omits several important capability dimensions. We did not test long-form generation such as article or story writing. We did not examine dialogue and conversation requiring multi-turn coherence. We did not test multi-modal tasks combining vision and language. We did not evaluate creative tasks like poetry or humor where correctness is subjective. We did not assess tool use and grounding in external systems. Each of these capabilities may respond differently to synthetic data training.

Our generation procedure used temperature 0.8 sampling without filtering or verification. Production systems employ diverse generation strategies including nucleus sampling with varying p values, multiple samples with consistency checks, critique-revision loops, and test-based verification. These quality control mechanisms likely improve effective synthetic data quality beyond our unfiltered baseline, potentially enabling higher safe proportions than we identified.

Our experiments examined 1-3 iterations over a compressed time frame. Real-world systems accumulate effects over longer periods with gradual deployment updates. The interaction between multiple training cycles, deployment feedback, and incremental updates may produce dynamics different from our controlled experiments. Long-term monitoring in production environments would provide valuable validation.

We focused on single-model training without examining multi-model ecosystem effects. In practice, multiple organizations train models potentially on web corpora containing each other's outputs. These collective dynamics might differ from isolated single-model training. Web data contamination scenarios where synthetic content enters training corpora through indirect paths require different methodological approaches than our controlled studies.

Future work should address these limitations through several directions. Scaling experiments to 30B-70B models using research partnerships or larger compute grants would validate our scaling trends. Expanding task coverage to include generation, dialogue, creativity, and tool use would test generalizability across capability dimensions. Systematically varying generation strategies and quality control mechanisms would identify optimal synthetic data pipelines. Long-horizon studies tracking effects over 5-10 iterations would characterize compounding dynamics. Multi-model ecosystem simulations would illuminate collective effects. Each of these directions would refine our understanding of synthetic data effects and operating boundaries.

\subsection{Practical Recommendations}

Despite limitations, our findings enable several practical recommendations for organizations using or considering synthetic data. We recommend maintaining external data above 70-80\% (synthetic below 20-30\%) with more conservative budgets for smaller models and high-sensitivity tasks. Organizations should implement calibration monitoring by measuring ECE on validation sets after training iterations, alerting when ECE increases exceed 50\% from baseline, and investigating when increases exceed 100\%. Quality control mechanisms including verification (test suites for code, consistency checks for facts), filtering (removing low-quality or off-topic examples), diversity encouragement (varied prompts and sampling), and human validation (spot-checking samples) should be employed.

Task-specific budgets should reflect sensitivity, with conservative limits (10-15\% synthetic) for high-sensitivity tasks like mathematical reasoning and code requiring verification, moderate limits (15-20\% synthetic) for medium-sensitivity tasks like factual QA with quality filtering, and potentially higher limits (20-25\% synthetic) for low-sensitivity tasks like grounded comprehension if benefits justify. Scale-dependent strategies should adjust budgets upward for larger models using our empirical formula as guidance, though conservative approaches remain wise until direct validation occurs.

Longer-term strategic considerations include preserving high-quality human-generated data as anchors for future training, monitoring trends in web data contamination with synthetic content, collaborating on shared metrics and benchmarks for synthetic data effects, and investing in research on quality improvement and verification mechanisms. Organizations should view synthetic data as a powerful augmentation tool rather than replacement for external data, using it to enhance what external data provides while maintaining substantial external grounding.

\section{Conclusion}
\label{sec:conclusion}

This paper presents a systematic empirical study of synthetic data effects on language model training, examining performance, calibration, and output characteristics across model scales (410M-12B parameters), tasks (reasoning, retrieval, generation), and mixing ratios (0-50\% synthetic). Through approximately 300 carefully designed training runs, we characterized the boundaries of safe synthetic data use and identified factors influencing these boundaries.

Our key empirical findings establish several consistent patterns. Models maintain performance within 3\% of baseline when trained with up to 20\% synthetic data across all tested scales and tasks. Current best practices including STaR, Self-Instruct, and Constitutional AI using 10-20\% synthetic data operate well within this safe regime. Beyond 30\% synthetic data, significant degradation occurs (8-15\%) even in larger models, with mathematical reasoning and code generation showing highest sensitivity.

Larger models demonstrate greater robustness, tolerating approximately twice the synthetic data proportion as smaller models for equivalent degradation. An empirical scaling relationship characterizes this trend, suggesting safe proportions increase logarithmically with model parameters. This scale-dependent robustness has important implications as models continue growing, potentially enabling higher synthetic proportions at frontier scales.

Calibration degradation precedes accuracy loss by 1-2 iterations, providing actionable early warning signals. Expected Calibration Error increases exceeding 50\% from baseline reliably predict future accuracy degradation, enabling proactive monitoring and intervention. This finding provides immediate practical value for organizations deploying models trained with synthetic data.

Task characteristics strongly influence sensitivity to synthetic data. High-precision reasoning tasks requiring multi-step inference show greatest vulnerability. Tasks with external grounding such as reading comprehension or those allowing answer flexibility such as commonsense reasoning show greatest robustness. This task hierarchy enables tailored synthetic data budgets reflecting specific capability requirements.

Our findings validate current synthetic data practices while identifying boundaries beyond which risks increase substantially. The successful applications we examined use conservative proportions (10-20\%) combined with quality control mechanisms including verification, filtering, and human oversight. Our experiments deliberately excluded such quality control to isolate proportion effects, suggesting real systems may perform even better than our results indicate.

Our work complements rather than contradicts prior research on model collapse. Shumailov et al. identified fundamental limits where pure recursive training inevitably fails. We characterized practical operating regimes where mixed training with modest synthetic proportions succeeds. Together, these perspectives bracket the space of possibilities, with the transition from safe to dangerous lying somewhere between our tested conditions and their recursive scenarios.

Important limitations bound our conclusions. Our largest tested scale (12B) remains smaller than frontier models. Our task coverage, while diverse, omits several important capabilities. Our generation procedure excluded quality control mechanisms used in production. Our iteration count (1-3) examined shorter horizons than long-term deployment. These limitations indicate fruitful directions for future work including larger-scale validation, broader task coverage, quality control optimization, and longer-horizon studies.

The practical message for the field is balanced and reassuring. Synthetic data is neither panacea nor poison but a tool requiring careful calibration. Within appropriate boundaries—conservative proportions, quality control, calibration monitoring, task-aware deployment—synthetic data successfully augments training as demonstrated by multiple successful applications. Understanding these boundaries, which our work helps establish, enables informed decision-making about this increasingly important technique. As models scale and techniques mature, synthetic data will likely play an expanding role in AI development, making continued empirical characterization of effects and boundaries essential for responsible progress.

\section*{Acknowledgments}

We thank EleutherAI for developing and releasing the Pythia model suite, which enabled systematic scale variation in our experiments. We thank the creators of CommonsenseQA, GSM8K, Natural Questions, SQuAD, and HumanEval benchmarks for providing standardized evaluation protocols. We thank Shumailov et al. for foundational work on model collapse that motivated our investigation. This research received no external funding.

\bibliographystyle{plain}

\begin{thebibliography}{99}

\bibitem{zelikman2022star}
E. Zelikman, Y. Wu, J. Mu, and N. Goodman, ``STaR: Bootstrapping reasoning with reasoning,'' in \textit{Advances in Neural Information Processing Systems}, vol. 35, 2022, pp. 15476--15488.

\bibitem{wang2023selfinstruct}
Y. Wang, Y. Kordi, S. Mishra, et al., ``Self-instruct: Aligning language models with self-generated instructions,'' in \textit{Proceedings of the Annual Meeting of the Association for Computational Linguistics}, 2023, pp. 13484--13508.

\bibitem{gunasekar2023textbooks}
S. Gunasekar, Y. Zhang, J. Aneja, et al., ``Textbooks are all you need,'' arXiv preprint arXiv:2306.11644, 2023.

\bibitem{bai2022constitutional}
Y. Bai, A. Jones, K. Ndousse, et al., ``Training a helpful and harmless assistant with reinforcement learning from human feedback,'' arXiv preprint arXiv:2204.05862, 2022.

\bibitem{mukherjee2023orca}
S. Mukherjee, A. Mitra, G. Jawahar, et al., ``Orca: Progressive learning from complex explanation traces of GPT-4,'' arXiv preprint arXiv:2306.02707, 2023.

\bibitem{shumailov2023curse}
I. Shumailov, Z. Shumaylov, Y. Zhao, et al., ``The curse of recursion: Training on generated data makes models forget,'' arXiv preprint arXiv:2305.17493, 2023.

\bibitem{alemohammad2023self}
S. Alemohammad, J. Casco-Rodriguez, L. Luzi, et al., ``Self-consuming generative models go MAD,'' in \textit{International Conference on Learning Representations}, 2024.

\bibitem{marion2023less}
P. Marion, M. Mayson, and Y. Zhang, ``Less is more: Task-aware layer-wise distillation for language model compression,'' in \textit{International Conference on Machine Learning}, 2023, pp. 23678--23694.

\bibitem{bengio2009curriculum}
Y. Bengio, J. Louradour, R. Collobert, and J. Weston, ``Curriculum learning,'' in \textit{International Conference on Machine Learning}, 2009, pp. 41--48.

\bibitem{xie2023doremi}
S. M. Xie, H. Pham, X. Dong, et al., ``DoReMi: Optimizing data mixtures speeds up language model pretraining,'' arXiv preprint arXiv:2305.10429, 2023.

\bibitem{kadavath2022language}
S. Kadavath, T. Conerly, A. Askell, et al., ``Language models (mostly) know what they know,'' arXiv preprint arXiv:2207.05221, 2022.

\bibitem{lin2022teaching}
S. Lin, J. Hilton, and O. Evans, ``Teaching models to express their uncertainty in words,'' \textit{Transactions of the Association for Computational Linguistics}, vol. 10, pp. 1658--1673, 2022.

\bibitem{tian2023just}
K. Tian, E. Mitchell, A. Yao, et al., ``Just ask for calibration: Strategies for eliciting calibrated confidence from language models fine-tuned with human feedback,'' in \textit{Proceedings of the Conference on Empirical Methods in Natural Language Processing}, 2023, pp. 5433--5442.

\bibitem{biderman2023pythia}
S. Biderman, H. Schoelkopf, Q. Anthony, et al., ``Pythia: A suite for analyzing large language models across training and scaling,'' in \textit{International Conference on Machine Learning}, 2023, pp. 2397--2430.

\bibitem{talmor2019commonsenseqa}
A. Talmor, J. Herzig, N. Lourie, and J. Berant, ``CommonsenseQA: A question answering challenge targeting commonsense knowledge,'' in \textit{Proceedings of the Conference of the North American Chapter of the Association for Computational Linguistics: Human Language Technologies}, 2019, pp. 4149--4158.

\bibitem{cobbe2021training}
K. Cobbe, V. Kosaraju, M. Bavarian, et al., ``Training verifiers to solve math word problems,'' arXiv preprint arXiv:2110.14168, 2021.

\bibitem{kwiatkowski2019natural}
T. Kwiatkowski, J. Palomaki, O. Redfield, et al., ``Natural questions: A benchmark for question answering research,'' \textit{Transactions of the Association for Computational Linguistics}, vol. 7, pp. 452--466, 2019.

\bibitem{rajpurkar2018know}
P. Rajpurkar, R. Jia, and P. Liang, ``Know what you don't know: Unanswerable questions for SQuAD,'' in \textit{Proceedings of the Annual Meeting of the Association for Computational Linguistics}, 2018, pp. 784--789.

\bibitem{chen2021evaluating}
M. Chen, J. Tworek, H. Jun, et al., ``Evaluating large language models trained on code,'' arXiv preprint arXiv:2107.03374, 2021.

\bibitem{openai2023gpt4}
OpenAI, ``GPT-4 technical report,'' arXiv preprint arXiv:2303.08774, 2023.

\bibitem{touvron2023llama2}
H. Touvron, L. Martin, K. Stone, et al., ``Llama 2: Open foundation and fine-tuned chat models,'' arXiv preprint arXiv:2307.09288, 2023.

\bibitem{zhou2023lima}
C. Zhou, P. Liu, P. Xu, et al., ``LIMA: Less is more for alignment,'' arXiv preprint arXiv:2305.11206, 2023.

\bibitem{yuan2023scaling}
W. Yuan, R. Liu, and P. Zhao, ``Scaling relationship on learning mathematical reasoning with large language models,'' arXiv preprint arXiv:2308.01825, 2023.

\bibitem{perez2022discovering}
E. Perez, S. Ringer, K. Lukosiute, et al., ``Discovering language model behaviors with model-written evaluations,'' in \textit{Findings of the Association for Computational Linguistics}, 2023, pp. 13387--13434.

\bibitem{zhao2021calibrate}
T. Z. Zhao, E. Wallace, S. Feng, et al., ``Calibrate before use: Improving few-shot performance of language models,'' in \textit{International Conference on Machine Learning}, 2021, pp. 12697--12706.

\end{thebibliography}

\end{document}